# ADAPTIVE FEDERATED LEARNING FOR SHIP DETECTION ACROSS DIVERSE SATELLITE IMAGERY SOURCES

*Tran-Vu La[1], Minh-Tan Pham[2], Yu Li[1], Patrick Matgen[1], and Marco Chini[1]*

[1]Luxembourg Institute of Science and Technology (LIST), 41 Rue du Brill, 4422 Sanem, Luxembourg

[2]Institut de Recherche en Informatique et Systèmes Aléatoires (IRISA), Université Bretagne Sud (UBS), 56000 Vannes, France

**ABSTRACT**

We investigate the application of Federated Learning (FL) for ship detection across diverse satellite datasets, offering a privacy-preserving solution that eliminates the need for data sharing or centralized collection. This approach is particularly advantageous for handling commercial satellite imagery or sensitive ship annotations. Four FL models—FedAvg, FedProx, FedOpt, and FedMedian—are evaluated and compared to a local training baseline, where the YOLOv8 ship detection model is independently trained on each dataset without sharing learned parameters. The results reveal that FL models substantially improve detection accuracy over training on smaller local datasets and achieve performance levels close to global training that uses all datasets during the training. Furthermore, the study underscores the importance of selecting appropriate FL configurations, such as the number of communication rounds and local training epochs, to optimize detection precision while maintaining computational efficiency.

## 1. INTRODUCTION

Over the last decade, Deep Learning (DL) has emerged as an essential tool for ship detection in satellite imagery, with most research adapting object detection models originally developed for computer vision (CV) tasks. These models are generally classified into one-stage detectors, like the YOLO series [1-2] and RetinaNet [3], and two-stage detectors based on the RCNN (Region-based Convolutional Neural Network) framework [4]. The choice between these approaches depends on application requirements, typically involving a trade-off between detection accuracy and processing speed: one-stage models offer faster predictions, while two-stage models provide greater accuracy. Most studies on DL-based ship detection have used high-resolution (HR) optical imagery because large datasets with this characteristic are freely available and can be used for training and testing without any extra annotation. Their high spatial resolution enables the detection of ships across various scales, from small vessels to large container ships. However, these datasets often lack critical information about specific regions of interest (ROIs), diverse ship categories, and varying cloud and met-ocean conditions. Conversely, lower-resolution (LR) satellite imagery (pixel sizes above 10 meters) has received less attention due to the scarcity of large, annotated datasets. As a result, finding small ships in LR datasets is still much harder for most DL models than it is for HR imagery [5-6], even though finding ships at different sizes is necessary for maritime surveillance.

However, the inconsistent performance of DL models across different datasets (HR and LR), acquired under various conditions and over diverse ROIs, poses a significant challenge. While some transfer learning techniques have been proposed to address this issue [7-8], these methods typically adapt to specific datasets rather than accommodating diverse ones. A potential solution is to gather a wide range of satellite imagery with different spatial and spectral resolution and train DL models on a combined more exhaustive dataset. However, this approach faces numerous challenges, including being time-intensive, costly, and encumbered by permissions and technical barriers, particularly when sharing commercial satellite images, an impractical option for operational workflows. Furthermore, confidentiality concerns often arise in ship detection over sensitive areas or under specific conditions, restricting the sharing of annotated datasets (e.g., those focusing on certain vessel types, regions, or scenarios). This limitation exacerbates the shortage of training data for some use cases.

To address the challenges of data collection and sharing in ship detection, this paper investigates the application of Federated Learning (FL) models for ship detection from multi-source satellite imagery. Unlike traditional DL approaches that require centralized data collection, FL enables decentralized model training while preserving data privacy. In an FL system, multiple clients train DL models on their local datasets and share only model updates with a central server. The server aggregates these updates and sends back a global model, ensuring that sensitive data remains on local devices. While FL has gained attention for Earth Observation (EO) applications such as land-use monitoring, land-cover classification [9], disaster management [10], environmental monitoring [11], and maritime transport (IoT) [12], its application to ship detection from multi-source satellite imagery has not been widely explored [13]. This study fills that gap by adapting several FL algorithms, including FedAvg [14], FedProx [15], FedOpt [16], and FedMedian [17], for ship detection across diverse optical datasets. To simulate real-world scenarios, we created eight datasets using satellite images from Geosat-2, Pléiades, Planet, Kompsat-2, RapidEye, Sentinel-2, Gaofen-1/6, and Landsat-8/9. These datasets, covering various ROIs and diverse cloud and met-ocean conditions, serve as individual clients in the FL network. The spatial resolutions of these datasets range from 0.75 m (Geosat-2) to 30 m (Landsat-8/9), making this study the first to comprehensively explore FL for ship detection across such a wide variety of remote sensing data. This work demonstrates the potential of FL to enable large-scale, privacy-preserving ship detection without requiring centralized data collection.

The paper is structured as follows: Section 2 covers data selection from optical satellite imagery. Section 3 outlines the methodology, including local, global, and FL-based training for ship detection. Section 4 presents experimental results, comparing training strategies, while Section 5 discusses findings and concludes the study.

## 2. DATASET PREPARATION

To implement FL models for ship detection, we construct eight datasets, each representing a client in the designed FL framework, as summarized in Table I. These datasets are grouped into HR and LR categories based on spatial resolutions below or above 10 meters, respectively. Among these, only Sentinel-2 and Landsat-8/9 datasets are publicly available, while the remaining datasets are commercial and have restricted access. Geosat-2, Pléiades, Kompsat-2, and RapidEye images are acquired in European maritime zones, while Planet, Sentinel-2, Gaofen-1/6, and Landsat-8/9 cover more diverse ROIs.

**Table I. Description of Datasets Used for FL-Based Ship Detection**

| No. Client | Dataset | Spatial resolution | Accessibility |
|---|---|---|---|
| 1 | Geosat-2 | 0.75 m | Commercial |
| 2 | Pléiades | 2 m | Commercial |
| 3 | Planet | 3 m | Commercial |
| 4 | Kompsat-2 | 4 m | Commercial |
| 5 | RapidEye | 5 m | Commercial |
| 6 | Sentinel-2 | 10 m | Public |
| 7 | Gaofen-1/6 | 16 m | Commercial |
| 8 | Landsat-8/9 | 30 m | Public |

Since most DL models for object detection are designed for RGB images (8 bits per channel), the original optical images are converted to this format and further divided into $512 \times 512$ pixel sub-images. We carefully select the datasets from various clients to enhance their diversity by including various cloud cover levels and met-ocean conditions, thereby ensuring a robust and comprehensive FL training environment.

Table II summarizes the data selection process for training FL models. Each client uses 200 sub-images for training and 67 sub-images for testing and validation of DL models for ship detection. Additionally, Group-All represents a combined dataset encompassing all eight datasets to evaluate the performance of DL models for the optimal solution.

**Table II. Data Selection for Training Federated Learning Models**

| Dataset | Total number | Training | Validation | Test |
|---|---|---|---|---|
| Each client | 334 | 200 | 67 | 67 |
| Group-All | 2672 | 1600 | 536 | 536 |

## 3. METHODOLOGY

To assess the performance of FL models for ship detection across eight datasets, we define three training scenarios, including

- Local training (Fig. 1): Each client independently trains DL models on its local dataset without sharing the trained models.
- Global training (Fig. 1): A single DL model is trained on the combined Group-All dataset, representing an ideal centralized training scenario.
- FL training (Fig. 2): Each client trains DL models locally and sends the trained models to the FL server for aggregation, enabling decentralized learning.

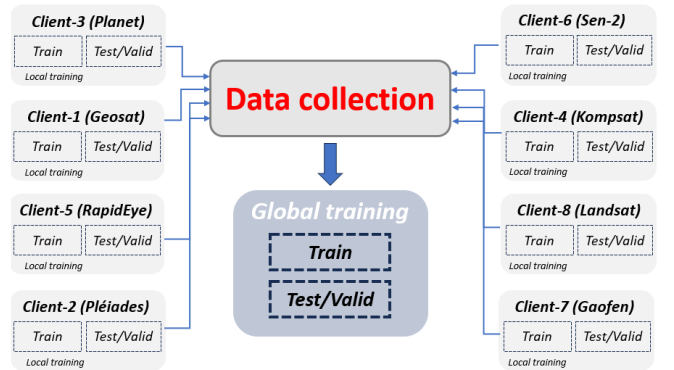

**Fig. 1.** Local and global training strategies for ship detection from multi-source satellite datasets: Local training without data sharing and global training with centralized data aggregation from various sources.

### 3.1. Local and Global Training

Fig. 1 illustrates the local and global training schemes, representing the baseline and ideal scenarios for ship detection from eight datasets. In the local training approach, each client independently trains a DL model using only its local dataset. The trained models are then evaluated on the combined Group-All dataset. This method can be considered as the least optimal training strategy due to the limited data diversity and lack of shared learning across datasets. In the global training strategy, data from all eight datasets are aggregated into a single combined dataset, simulating a centralized training environment. The global DL model is trained on this comprehensive dataset, benefiting from the full diversity of ship types, environmental conditions, and spatial resolutions. As such, this approach is considered the optimal training strategy for ship detection due to its broad data representation and enhanced generalization potential.

### 3.2. Federated Learning Models

Fig. 2 illustrates the proposed FL framework for ship detection, consisting of a central server and eight clients ($K_{i=1-8}$), each representing a unique satellite dataset. The process begins with the server distributing initial model weights ($w_0$) to all clients. Each client trains the model on its local dataset ($K_i$) for $E$ epochs, optimizing an objective function ($F_{k,i}$) to produce updated model parameters ($w_{k,i}$). These updated $w_{k,i}$ are then sent back to the server, which aggregates them using a specific aggregation function ($f_a$), determined by the FL algorithm employed. This training and aggregation cycle is repeated over $T$ rounds until the model's performance converges.

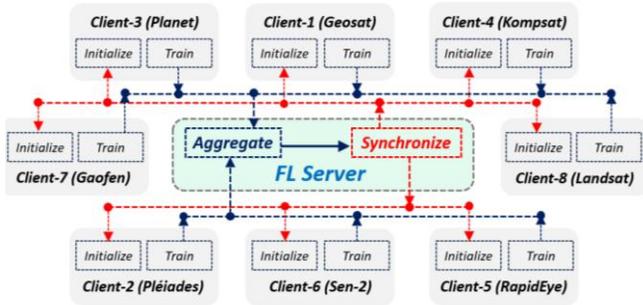

**Fig. 2.** Federated Learning framework for ship detection, consisting of a central server and eight clients, each representing a distinct satellite dataset.

In this study, we implement four FL strategies: FedAvg [13], FedProx [14], FedOpt [15], and FedMedian [16]. FedAvg and FedProx share a similar $f_a$, computing the average of $w_{k,i}$. However, they differ in the functions $F_{k,i}$. FedProx introduces a proximal term that scales the local updates, constraining them closer to the initial global model, thus mitigating local dataset biases. In contrast, FedMedian retains the $F_{k,i}$ as that of FedAvg but changes the $f_a$ to compute the median of client-updated weights, enhancing robustness against outliers. FedOpt applies an advanced $f_a$ using multiple hyperparameters, improving model convergence and communication efficiency. Detailed descriptions and algorithms for these FL models can be found in [13-16].

### 3.3. Deep Learning Models for Ship Detection and Evaluation Metrics

We have adopted the YOLOv8 model [18] for training on each client's local dataset as well as the combined Group-All dataset. As a state-of-the-art object detection model in CV, YOLOv8 features a modular and scalable architecture, making it adaptable to various platforms. Its configurable versions and backbones enable efficient feature extraction from input images, ensuring robust object detection across different scales and object categories.

To evaluate the performance of the proposed training strategies (local, global, and FL), we employ widely used object/ship detection metrics, including Recall, Precision, and mean Average Precision at 50% Intersection over Union (mAP50), as recommended in prior studies [1-6].

## 4. EXPERIMENTAL RESULTS

To evaluate the performance of FL models for ship detection, we compare the mAP scores obtained using FedAvg, FedProx, FedOpt, and FedMedian against local and global training approaches. To ensure a fair comparison across all training strategies, we maintain the same total number of training epochs, $E_{total} = 150$, for each method. For local training, each client trains its model independently on its local dataset for 150 epochs. In global training, the combined Group-All dataset is used to train the model for 150 epochs. In the FL training strategy, we set the number of communication rounds $T = 10$ and the number of local training epochs per client $E = 15$, ensuring the total number of training epochs follows $E_{total} = T \times E$. This setup allows a fair comparison of model performance across the different training schemes.

### 4.1. Performance of Various FL Models for Ship Detection of Eight Clients

Fig. 3 illustrates the mAP scores for all training strategies evaluated over 150 epochs, ensuring model convergence. Global training on the combined Group-All dataset achieves the highest mAP score of 0.843, while local training yields the lowest average performance of 0.545, based on the YOLOv8 results across the eight clients. Among the FL models, FedAvg, FedProx, and FedOpt demonstrate comparable mAP scores, with FedMedian performing slightly lower. Although the FL models underperform compared to global training (0.743 vs. 0.843), they still significantly outperform local training (0.743 vs. 0.545). Additionally, the computation times for all four FL models are nearly identical, likely due to the simplicity of detecting a single class (ship) across all datasets.

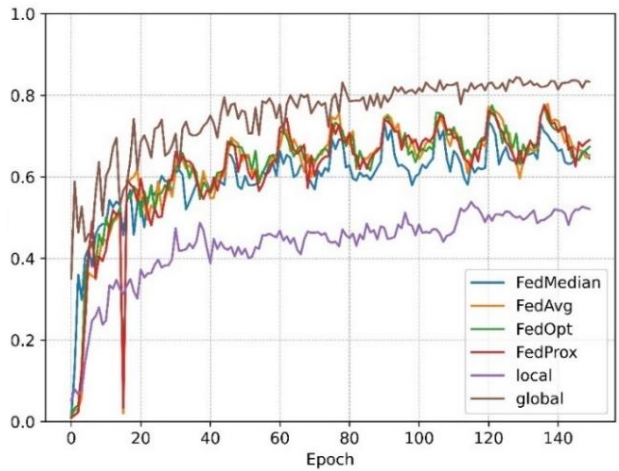

**Fig. 3.** Performance of local and global training strategies (Fig. 1) and various FL models trained with eight clients (Fig. 2).

### 4.2. Comparison of Different FL Training Strategies

In ship/object detection, the performance of FL models is assessed not only by mAP scores but also by calculation time and data-sharing security. While the latter falls outside the scope of this study, calculation time is a crucial factor influenced by the number of communication rounds ($T$) between the FL server and clients. Determining the optimal value of $T$ is challenging, as it depends on several factors, including the specific application (e.g., object detection, classification, IoT), desired accuracy, and acceptable computation time. Currently, no precise method exists for selecting the optimal $T$, making this a critical consideration in FL model deployment. In this study, to ensure a fair comparison across all training strategies, we set $E_{total} = 150$ and propose four configurations of $T$ and $E$ (local training epochs) to evaluate the performance of FL models:

- *Opt-1*: $T = 3$ rounds × $E = 50$ epochs
- *Opt-2*: $T = 5$ rounds × $E = 30$ epochs
- *Opt-3*: $T = 10$ rounds × $E = 15$ epochs
- *Opt-4*: $T = 15$ rounds × $E = 10$ epochs

Table III shows the highest mAP scores achieved by FedAvg for ship detection across eight clients/datasets under four tested scenarios (*Opt-1/4*). We focus on FedAvg since all four FL models yielded similar mAP scores, as seen in Fig. 3. *Opt-2*, *Opt-3*, and *Opt-4* provide comparable mAP scores of approximately 0.74, while *Opt-1* underperforms slightly at 0.69. Nevertheless, all four configurations outperform local training (0.54).

Additionally, the mAP scores across the eight clients are relatively consistent, likely due to the balanced datasets used. In terms of calculation time, it increases with the number of $T$. As $T$ rises from 3 to 5, 10, and 15, the FL model's performance improves. However, the improvement in mAP scores between $T = 5$, 10, and 15 is marginal, while the calculation time increases significantly. This finding suggests that selecting an appropriate value for $T$ is crucial, balancing model accuracy and computational efficiency.

**Table III.** Performance of FedAvg FL model trained with different strategies of rounds and epochs

| Training Strategy | Opt-1<br>3 × 50 | Opt-2<br>5 × 30 | Opt-3<br>10 × 15 | Opt-4<br>15 × 10 |
|---|---|---|---|---|
| Geosat-2 | 0.725 | 0.746 | 0.777 | 0.781 |
| Pléiades | 0.680 | 0.732 | 0.696 | 0.739 |
| Planet | 0.706 | 0.744 | 0.746 | 0.769 |
| Kompsat | 0.708 | 0.743 | 0.745 | 0.778 |
| RapidEye | 0.658 | 0.714 | 0.710 | 0.698 |
| Sentinel-2 | 0.704 | 0.723 | 0.742 | 0.729 |
| Gaofen | 0.649 | 0.687 | 0.770 | 0.739 |
| Landsat | 0.706 | 0.735 | 0.756 | 0.753 |
| Avg. score | 0.692 | 0.728 | 0.743 | 0.748 |
| Local | 0.545 | | | |
| Global | 0.843 | | | |
| Cal. time | 6261 | 9005 | 15951 | 22700 |

### 5. CONCLUSION

This paper highlights the advantages of applying Federated Learning (FL) for ship detection from multi-source satellite datasets without the need for data sharing or centralized collection. This approach addresses challenges related to data permissions, technical barriers, and time-consuming data aggregation. While FL models yield lower mAP scores than global training, they significantly outperform local training methods. Our findings also emphasize the importance of selecting an optimal FL configuration, balancing the number of communication rounds, local epochs, and model selection to achieve a trade-off between detection accuracy and computational efficiency.

Future work will explore FL performance under more complex scenarios, including clients with unbalanced datasets and datasets with distinct features, such as the combination of radar and optical satellite imagery.